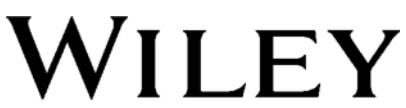

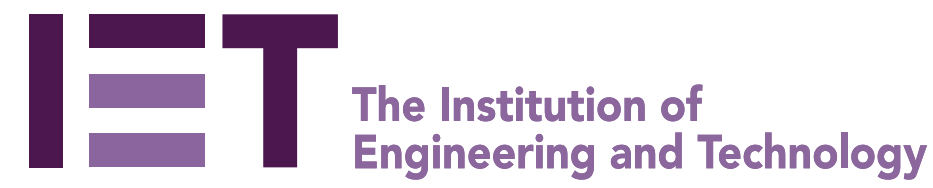

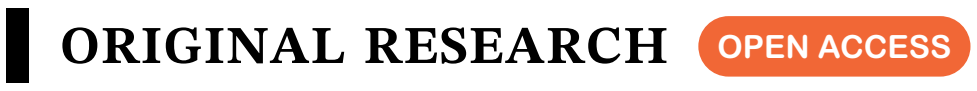

# Designing Cellular Manufacturing Systems in the Presence of Alternative Process Plans

Md. Kutub Uddin[1] | Md. Saiful Islam[2] | Md Abrar Jahin[3] | Md. Tanjid Hossen Irfan[2] | Md. Saiful Islam Seam[2] | M. F. Mridha[4]

[1]Department of Mechanical Engineering, Khulna University of Engineering & Technology, Khulna, Bangladesh | [2]Department of Industrial Engineering and Management, Khulna University of Engineering & Technology, Khulna, Bangladesh | [3]Thomas Lord Department of Computer Science, Viterbi School of Engineering, University of Southern California, Los Angeles, California, USA | [4]Department of Computer Science, American International University–Bangladesh, Dhaka, Bangladesh

**Correspondence:** M. F. Mridha (firoz.mridha@aiub.edu)

## ABSTRACT

In the design of cellular manufacturing systems (CMS), numerous technological and managerial decisions must be made at both the design and operational stages. The first step in designing a CMS involves grouping parts and machines. In this paper, four integer programming formulations are presented for grouping parts and machines in a CMS at both the design and operational levels for a generalised grouping problem, where each part has more than one process plan and each operation of a process plan can be performed on more than one machine. These four integer programming formulations have the ability to simultaneously optimise cell formation and operational routing decisions within a unified framework, whereas prior models often treat these as sequential or independent stages. By integrating machine flexibility and alternative process plans directly into the mathematical objective, this approach ensures a more globally optimal configuration for generalised grouping problems. Minimising inter-cell and intra-cell movements is achieved by assigning as many consecutive operations of a part type as possible to the same cell and machine. This objective is evaluated against alternatives like reducing machine investment and operating costs. Numerical examples demonstrate how the proposed formulations work and highlight their effectiveness in practice.

## 1 | Introduction

An increasingly competitive environment and ever-changing customer preferences have compelled manufacturers to develop a manufacturing system that can support a large variety of high-quality products, small lot sizes and flexibility in terms of frequent product changes, all while minimising manufacturing costs and lead times as much as possible. A conventional mass production system, which utilises special-purpose machines arranged in a line to facilitate smooth production, is characterised by low in-process inventory, high efficiency and simplified management control. However, it is unsuitable for handling a multi-product, small-lot-size situation [1]. On the other hand, batch and job production, which involves a large number of general-purpose machines, adopts a functional layout to suit the economics of medium and low-volume production with moderate and high variety. However, in such batch and job manufacturing systems, a part may spend up to 95% of the total production time in waiting and travelling, leading to long and uncertain throughput times [2]. Thus, flexibility is acquired at the cost of productivity.

Therefore, the main concerns in the emerging business world are flexibility in terms of frequent product changes and smaller batch sizes, shorter throughput times and higher productivity. Cellular Manufacturing Systems (CMS), which are based on the philosophy of Group Technology (GT), have been recognized as one of the recent technological innovations in job shop or batch-type production systems. They aim to achieve economic advantages similar to those of mass production systems while maintaining the flexibility of job or batch production systems [3]. The idea behind a cellular manufacturing system is to partition or decompose the entire production system into small autonomous subsystems and identify subsets of parts with similar design or processing requirements so that a single subsystem can process each subset of parts. In the context of GT, each such subsystem is termed a machine cell and each subset of parts is referred to as a part family. The identification of machine cells and part families, known as machine-component grouping, is the first step in designing a cellular manufacturing system.

The grouping problem can arise in two different situations. First, due to changes in market demand, some new products may need to be introduced, some existing products may need to be discontinued or both [4]. In such a real-life scenario, acquiring new machines, removing existing machines or both may be necessary. It is also possible that the existing set of machines is necessary and sufficient for the new range of parts. Therefore, the existing manufacturing system (with or without the addition of new machines) must be reorganised into a new cellular structure. This problem of reorganization and regrouping of machines can be considered an operational-level problem. Second, there is the situation of setting up a new cellular manufacturing system to produce a set of new parts. In this case, the problem is to determine the requirements for machines of various types and to identify the machine cells and part families [5]. This problem is clearly a design-level problem.

The main objective of this study is to develop mathematical models for addressing the CMS design problem in the presence of alternative process routes. Specifically, the study aims to achieve the following objectives: (i) To develop four distinct integer programming formulations for the design and operational management of CMS. (ii) To address the generalised grouping problem by integrating alternative process plans and machine flexibility into a unified optimization framework. (iii) To minimise total inter-cell and intra-cell material handling movements, providing a robust decision-support tool for manufacturing practitioners.

The paper is structured as follows: Section 2 reviews the relevant literature. In Section 3, we develop the necessary notations and generate formulations for several grouping scenarios. Section 4 illustrates the application of these models with numerical examples and the analysis of the results is also presented in this section. Conclusions are provided in Section 5.

## 2 | Literature Review

Several methods for forming machine-component cells have been developed over the years. Almasarwah and Süer [6], Danilovic and Ilic [7], Funke and Becker [8] and Gauss et al. [9] have all published comprehensive assessments of these approaches. These approaches are broadly classified into two categories: (i) Methods for determining part families according to parameters such as tolerance, material and geometry [10] and (ii) The production process is directly analysed using Production Flow Analysis (PFA) techniques, particularly part process plans or routings [11]. A significant number of existing approaches utilise the PFA (Production Flow Analysis) approach, assuming that each part has a single process plan and that each operation in a process plan can be carried out on only one machine. In practice, however, a part may have multiple process plans and each operation within a process plan could be performed on various machines. The grouping problem in such cases involves determining machine cells and process plan families in such a way that each machine cell can execute at least one process plan family, assigning operations to machines and selecting a single process plan for each part.

Alimian et al. [12] proposed a new integer nonlinear programming approach for dynamic cellular manufacturing systems with worker assignment. This model can estimate the optimal labour assignments, cell designs and process plans for each part type at every stage of the planning horizon. The importance of considering human factors in cellular manufacturing (CM) cannot be overstated, as neglecting them may significantly reduce the benefits of CM. Cell formation is one of the main concerns in planning CMS. Motahari et al. [10] presented an optimization method using simulated annealing (SA) to address this issue. The methodology they used comprises three key elements: identifying optimal process routings, classifying machines into cell groups and classifying parts into part families. This proposed method has been contrasted with a genetic algorithm (GA)-based methodology using a numerical example from the broader field. Mathematical programming is commonly used for modelling CMS problems, with typical goals including reducing material handling costs between cells or maximising the total sum of parts' similarity within each cell.

For CMS design, Bortolini et al. [13] proposed an extensive mathematical programming approach, demonstrating the significance of addressing several design considerations in an integrated manner. To create a robust cellular manufacturing system (RCMS) that accounts for dynamic production and multi-period production planning, Alimian et al. [12] suggested a novel integrated mathematical model. This model allows for flexibility in production planning (production/subcontracting) by designing product mixes within the limited capacity at each planning horizon without altering the manufacturing cell structure, making the problem more realistic and feasible. Salimpour et al. [14] developed an effective optimization strategy that accounts for part processing time uncertainty to solve a production planning problem using a dynamic cellular manufacturing system (DCMS). This approach utilises a new integrated mathematical model. In real-world scenarios, this type of uncertainty must be considered when evaluating the manufacturing system's efficiency. Additionally, certain real-world production features have been considered, such as operator assignment, cell arrangement within a cell, machine capacity, machine relocation and reliability and alternative process routes.

Furthermore, Kesavan et al. [15] proposed a mathematical framework for a Cell Formation Problem (CFP) based on the concept of CMS cell utilization. This method can optimise the cell layout by simultaneously reducing the number of exceptional elements (EEs) and voids in the cells. The mathematical model provides an effective solution using a genetic algorithm. To build a structural architecture for a DCMS, Pérez-Gosende et al. [16] suggested a unique mixed-integer nonlinear programming approach. Klausnitzer and Lasch [17] employed search methods with linear programming models to create a cell layout and channel flow. Vidal et al. [18] provided a two-step process for creating a CMS cell layout that reduces lead time for manufactured parts. Based on the related clustering-layout problem by Golmohammadi et al. [19], the bidirectional linear flow layout is the suggested approach for arranging machine cells to minimise inter-cell flow costs. A three-phase approach based on the cut tree network model is created for this mixed situation.

One of the grouping objectives considered by some of the above works is maximising the similarity coefficient among the members within a family or cell. However, a significant disadvantage of the similarity coefficient objective function is that it may fail to produce disjoint groups even when they exist. To the best of our knowledge, none of the aforementioned methods specifically consider minimising motions both within and between cells as an objective function for solving the grouping problem. Furthermore, the problem is often addressed in stages due to the hierarchical nature of most algorithms and methods and the optimal values at each stage can potentially lead to suboptimal overall grouping performance. Therefore, this study aims to optimise the cellular manufacturing system by minimising both intra-cell and inter-cell movements, which are major contributors to material handling costs and production inefficiencies. The major design-level objectives (setting up a new cellular system) considered are (i) minimisation of total investment cost while keeping inter-cell movement at zero, (ii) minimisation of the amortised cost of machines plus total operating costs while keeping inter-cell movement at zero and (iii) minimisation of both inter-cell and intra-cell movement while keeping total investment costs within budgetary limits. The suitability of these objectives is also discussed with examples.

## 3 | Mathematical Models

### 3.1 | Notation

The following notation is used for the development of mathematical models.

#### *3.1.1 | Indices*

$k$ = part

$m$ = machine type

$p$ = process plan/process route

$i, j, s$ = operation

$f$ = part family

$c$ = machine cell

#### *3.1.2 | Parameters*

$K$ = total number of parts.

$M$ = total number of machine types.

$S$ = total number of operations.

$\mathrm{PP}(k)$ = set of process plans for part $k$.

$\mathrm{TPP}(k)$ = total number of process plans for part $k = |\mathrm{PP}(k)|$

$\mathrm{PR}(k)$ = set of process routes for part $k$.

$\mathrm{TPR}(k)$ = total number of process routes for part $k = |\mathrm{PR}(k)|$

$S(kp)$ = set of operations in process plan $p$ of part $k$.

$\mathrm{TS}(kp)$ = total number of operations in process plan $p$ of part $k = |S(kp)|$.

$\mathrm{SUC}(s)$ = set of operations that can succeed an operation $s$ in any process plan. In case of a strict technological sequence of operations, the number of elements in the set will be $l$ and when there is no technological sequence at all, the number of elements in the set will be equal to $\mathrm{TS}(kp) - 1$ $\mathrm{O}(kp)$ = set of pairs of consecutive operations for process plan $p$ of part $k$

$$= \{(i, j); i, j \in \mathrm{S}(kp), i < j, j \in \mathrm{SUC}(i)\}.$$

In defining the set of pairs of consecutive operations for a process plan, care has been taken so that (i) only the pairs of logical consecutive operations become the members and (ii) these pairs are taken only once, ignoring the order between the two operations in a pair. This definition works well for both strict and flexible technological sequences of operations.

$\mathrm{TO}(kp)$ = total number of pairs of consecutive operations to be considered for process plan $p$ of part $k = |\mathrm{O}(kp)|$

$M(s)$ = set of machine types that can perform operation $s$

$\mathrm{TM}(s)$ = total number of machine types that can perform operation $s = |M(s)|$

$M(i) \cap M(j)$ = set of machine types that can perform both operations $i$ and $j$

$R(kp)$ = set of machine types required to process part $k$ using process route $p$

$R(k)$ = set of machine types required to process part $k$, in case part $k$ has only 1 process route (simple grouping problem)

$\mathrm{TR}(kp)$ = total number of machine types required to process part $k$ using process route $p = |R(kp)|$

$\mathrm{TR}(k)$ = total number of machine types required to process part $k$ in case part $k$ has only 1 process route (simple grouping) $|R(k)|$

$F$ = total number of part families

$C$ = total number of machine cells

$OC_{s(kp)m}$ = operating cost for machine $m$ performing operation $s$ of process plan $p$ of part $k$

$OC_{kp}$ = total operating cost of a process route $p$ of part $k$ = the sum of individual operating costs of all its operations on the corresponding machines

$t_{s(kp)m}$ = time taken by machine $m$ to perform operation $s$ of process plan $p$ of part $k$

$t_{kpm}$ = time required on machine $m$ to produce one unit of part $k$ using process route $p$

$d_k$ = demand for part $k$

$b_m$ = time available on each machine of type $m$

$A_m$ = number of copies available of machine type $m$

$\max_c$ = maximum number of machines allowed in cell $c$

$\min_c$ = minimum number of machines to be assigned to cell $c$

TOC = budget limit on total operating cost.

$I_m$ = investment cost per machine of type $m$

$I_{\mathrm{am}}$ = amortised cost per machine of type $m$

$B$ = budget limit on total investment on machines.

#### 3.1.3 | *Decision Variables*

$$X_{s(kp)mc} = \begin{cases} 1 & \text{if operation } s \text{ in process plan } p \text{ of part } k \text{ is assigned to machine } m \text{ in cell } c \\ 0 & \text{otherwise} \end{cases}$$

$$Y_{kp} = \begin{cases} 1 & \text{if process plan/process route } p \text{ is selected for part } k \\ 0 & \text{otherwise} \end{cases}$$

$$Y_{kpc} = \begin{cases} 1 & \text{if process plan } p \text{ is selected or part } k \text{ and part } k \text{ must visil cell } c \text{ for processing} \\ 0 & \text{otherwise} \end{cases}$$

$$Q_{kc} = \begin{cases} 1 & \text{if part } k \text{ visits cell } c \text{ or processing at some stage (for simple grouping situation)} \\ 0 & \text{otherwise} \end{cases}$$

$$Z_{mc} = \begin{cases} 1 & \text{if machine } m \text{ is assigned to cell } c \\ 0 & \text{otherwise} \end{cases}$$

$$\beta_{mc} = \begin{cases} 1 & \text{if part } k \text{ belongs to part family corresponding to cell } c \\ 0 & \text{otherwise} \end{cases}$$

$N_{mc}$ = total number of machines of type $m$ assigned to cell $c$

### 3.2 | Model Assumptions

To formulate the CMS design problem in the presence of alternative process plans, the following assumptions are adopted. These assumptions are consistent with common practices in the CMS literature and ensure tractability of the proposed optimization formulations. (i) The demand for each part type is deterministic and remains constant throughout the planning horizon. (ii) Processing times for each operation on every compatible machine type are known and deterministic. (iii) Machines are assumed to be available at all times during the production period; stochastic factors such as machine breakdowns and maintenance schedules are not explicitly modelled. (iv) Each part follows exactly one process route selected from its set of alternative process plans and each operation within that route is assigned to precisely one machine in one cell. (v) Material handling movements (inter-cell or intra-cell) are generated only when consecutive operations are performed on different machines or in different cells. (vi) Machine cell sizes are restricted within predefined minimum and maximum capacities to ensure manageable control and resource utilization. (vii) Setup times are either negligible or integrated into the processing times for each operation.

### 3.3 | Grouping at Operational Level: Reorganization of an Existing System

#### 3.3.1 | *Formulation I: Minimisation of Intercell and Intracell Movements*

##### 3.3.1.1 | *Objective Function.*

The objective here is to minimise the total intercell and intracell movement of parts. These movements are generated in three ways, depending on the assignment of any set of consecutive operations. First, consecutive operations are assigned to the same machine in the same cell, so there is neither intercell nor intracell movement. Second, consecutive operations are assigned to different cells, causing one intercell movement. Finally, consecutive operations are assigned to different machines in the same cell, giving rise to one intracell movement.

From this analysis, it is clear that the minimisation of intercell movements is equivalent to the maximisation of the assignment of consecutive operations to the same cell and that the minimisation of intracell movements is equivalent to the maximisation of the assignment of consecutive operations to the same machine in the same cell. As a result, the objective function is the count of instances where consecutive operations are assigned to the same cell or to the same machine. Thus, the objective function can be written as:

$$\max\left[\sum_{c=1}^{C}\sum_{k=1}^{K}\sum_{p\in \mathrm{PP}(k)}\sum_{(i,j)\in \mathrm{O}(kp)}\sum_{m\in M(i),n\in M(j),m\neq n} X_{i(kp)mc}.X_{j(kp)nc} + \sum_{c=1}^{C}\sum_{k=1}^{K}\sum_{p\in \mathrm{PP}(k)}\sum_{(i,j)\in \mathrm{O}(kp)}\sum_{m\in M(i)\cap M(j)} X_{i(kp)mc}.X_{j(kp)nc}\right]. \tag{1}$$

The first term in the objective function (1), considering the assignment of consecutive operations to the same cell, represents the weighted intercell movements. The weighted intracell movements are represented by the second term, which considers the assignment of consecutive operations to the same machine in the same cell. The above formulation can be used for both situations: where there is a strict technological sequence among the operations or where there is no technological sequence specified. For example, let process plan $p$ of part $k$ have three operations, 1, 2 and 3, in the case of a strict technological sequence. Let the operations be performed in a natural sequence. Then, for intercell movements, there will be two product terms, viz., $X_{1(kp)mc}\cdot X_{2(kp)nc}$ and $X_{2(kp)mc}\cdot X_{3(kp)nc}$. These terms have to be expanded for all feasible machines and cells. On the other hand, if there is no prespecified technological sequence, then there will be three product terms: $X_{1(kp)mc}\cdot X_{2(kp)nc}$, $X_{1(kp)mc}\cdot X_{3(kp)nc}$ and $X_{2(kp)mc}\cdot X_{3(kp)nc}$ which again, have to be expanded for all feasible machines and cells. From the above terms, it is obvious that in the presence of a strict technological sequence, assigning the first and the third operations in the same cell and the second operation in some other cell will have the same number of intercell movements as assigning the three operations in three different cells.

The objective function given in Equation (1) contains a large number of quadratic terms and thus is computationally complex. One of the popular approaches for solving such problems is to linearise the quadratic terms. Glover and Woolsey [20] suggested a methodology to linearise the quadratic terms at the cost of additional continuous variables and some constraints. The method is as follows. Replace each quadratic term with a continuous variable and three additional constraints. For example, the quadratic term $X_i.X_j$ can be replaced by a continuous variable $V_{ij}$ and the following three additional constraints:

$$X_i \geq V_{ij}, \tag{1a}$$

$$X_j \geq V_{ij}, \tag{1b}$$

$$X_i + X_j - V_{ij} \leq 1. \tag{1c}$$

Modi & Shanker [21] reported that the third constraint, namely, Equation (1c) is redundant in case both the variables are binary and then constraints (1a) and (1b) are sufficient to linearise a quadratic term involving two binary variables.

*3.3.1.2* | *Constraints.*

1. Selection of one process plan for each part: This constraint, by summing the variable $Y_{kp}$ over the whole set of process plans available for a part $k$ and by equalising this sum to 1, ensures that one and only one process plan is selected for each part.

$$\sum_{p\in \mathrm{PP}(k)} Y_{kp} = 1 \quad k = 1, \ldots, K. \tag{2}$$

2. Assignment of operations: An operation of a selected process plan is assigned to one and only one machine out of several alternative machines. These constraints do not bind the assignment of all the operations of a selected process plan to machines belonging to one cell.

$$\sum_{c=1}^{C}\sum_{m\in M(s)} X_{s(kp)mc} = Y_{kp} \quad k = 1, \ldots, K; \forall\, p \in \mathrm{PP}(k);\ \forall\, s \in S(kp). \tag{3}$$

3. Loads on machines: An operation is assigned to a machine type $m$ in cell $c$ only if cell $c$ has a non-zero number of copies of machine type $m$. Moreover, the total time required to process all operations that are assigned to a machine type in a cell must not exceed the cumulative capacity of that machine type in that cell:

$$\sum_{k=1}^{K}\sum_{p\in \mathrm{PP(k)}}\sum_{s\in S(kp)} d_k.X_{s(kp)mc}.t_{s(kp)m} \leq b_m.N_{mc}m = 1, \ldots, M;\ c = 1, \ldots, C. \tag{4}$$

4. Number of machines assigned to a cell: For proper control and supervision of the cell and for effective utilization of the resources, it is advisable to keep the number of machines in a cell within a certain range. Small groups are undesirable, even if disjointed, because they may lead to the underutilisation of resources. On the other hand, assigning too many machines to a cell may lead to a situation where proper control and supervision become difficult. Generally, small group size causes increased intercell movement, while large group size tends to have more intracellular movements. Two extreme examples will clarify this point. Machine groups with a single machine in each will have no intracell movement but a lot of intercell movements, whereas a large group consisting of all the machines in the system will have no intercell movements but many intracell movements. Both these cases go against the concept of group technology because the philosophy of GT is to assign machines to cells in a judicious manner so that both intracell and intercell movements and, consequently, material handling activities are minimised. Thus, for better utilization of resources and proper control and supervision, the upper and lower limits on the number of machines in a cell must be specified.

$$\sum_{m=1}^{M} N_{mc} \geq \min{}_c \quad c = 1, \ldots, C, \tag{5a}$$

$$\sum_{m=1}^{M} N_{mc} \geq \max_{c} \quad c = 1, \ldots, C. \tag{5b}$$

5. Number of copies available of each machine type: The total number of copies of each machine type assigned to various cells should not exceed the available number of copies.

$$\sum_{c=1}^{C} N_{mc} \leq A_{m} \quad m = 1, \ldots, M. \tag{6}$$

*3.3.1.3* | *Grouping at the Design Level: Setting up a New System.* In this subsection, some of the existing models that deal with the design of a cellular manufacturing system will be discussed and subsequently, the modified forms of these models will be presented.

### *3.3.2* | *Formulation II: Minimisation of Investment Cost on Machines*

This formulation is a modification to the model of Rajamani et al. [22]. They presented an integer programming model to solve the design-level grouping problem in which the total intercell movement is kept at zero (forcing the formation of strictly disjoint cells) and the objective is to minimise the total investment costs. The model includes a nonlinear constraint on the capacity of each machine type in each cell and linear constraints regarding the selection of a process plan for each part, assignment of operations for the selected process plans to machines, a budgetary limit on operating costs and cell size. A modification is made to their model so as to make all constraints linear. The formulation is as follows:

*3.3.2.1* | *Objective Function.* The objective function is to minimise the total investment cost on machines simply:

$$\min \sum_{c=1}^{C} \sum_{m=1}^{M} I_{m} \,.\, N_{mc}. \tag{7}$$

*3.3.2.2* | *Constraints.*

1. Selection of only one process plan for each part: This constraint is similar to constraint (2), except that in the earlier case, there was no binding to assign the selected process plan to only one cell for its processing. The present constraint will assign the selected process plan to only one cell for its total processing:

$$\sum_{c=1}^{C} \sum_{p \in \mathrm{PP}(k)} Y_{kpc} = 1 \quad k = 1, \ldots, K. \tag{8}$$

2. Assignment of operations: This constraint is similar to constraint (3) in the sense that each operation is assigned to a unique machine. The difference between the two is that constraint (3) does not assign all the operations of the selected process plan to machines belonging to the same cell, whereas the following constraint will assign all the operations of a selected process plan to machines in the particular cell $C$ for which $Y_{kpc}$ is one.

$$\sum_{m \in M(s)} X_{s(kp)mc} = Y_{kpc} \; c = 1, \ldots, C; \; k = 1, \ldots, K; \quad \forall\, p \in \mathrm{PP}(k); \; \forall\, s \in S(kp). \tag{9}$$

3. Total operating cost: Performing an operation on a particular machine involves some operating costs. The following constraint will assign the operations of selected process plans to machines in such a way as to keep the total operating cost within prespecified limits.

$$\sum_{c=1}^{C} \sum_{k=1}^{K} \sum_{p \in \mathrm{PP}(k)} \sum_{s \in S(kp)} \sum_{m \in M(s)} d_{k}.X_{s(kp)mc} \cdot \mathrm{OC}_{s(kp)m} \leq \mathrm{TOC}. \tag{10}$$

4. Loads on machines: This constraint is the same as Equation (4) and is duplicated here:

$$\sum_{k=1}^{K} \sum_{p \in \mathrm{PP}(k)} \sum_{s \in S(kp)} d_{k}.X_{s(kp)mc}.t_{s(kp)m} \leq b_{m} \cdot N_{mc} \quad m = 1, \ldots, M; \; c = 1, \ldots, C. \tag{11}$$

5. Number of machines assigned to a cell: To be consistent with Rajamani et al.'s model, only the upper limit on cell size is considered here, which is the same as constraint (5b) and is duplicated here:

$$\sum_{m=1}^{M} N_{mc} \leq \max_{c} \quad c = 1, \ldots, C. \tag{12}$$

It would be interesting to see how this formulation compares with Rajamani et al.'s [22] model, which is a modification. Below are the number of binary variables, integer variables, continuous variables and constraints for formulation II and Rajamani et al.'s [22] model (after linearisation).

| **Present model** | |
|---|---|
| No. of binary variables | $C\left[\sum_{k=1}^{K} \mathrm{TPP}(k) + \sum_{k=1}^{K} \sum_{p \in \mathrm{PP(k)}} \sum_{s \in S(kp)} \mathrm{TM}(s)\right]$ |
| No. of integer variables | $M{\cdot}C$ |
| No. of continuous variables | 0 |
| No. of constraints | $K + C\sum_{k=1}^{K} \mathrm{TPP}(kp) + 1 + M \cdot C + C$ |

| Model of Rajamani et al. [22] (Linearised) | |
|---|---|
| No. of binary variables | $\sum_{k=1}^{K} \mathrm{TPP}(k) + \sum_{k=1}^{K} \sum_{p \in \mathrm{PP(k)}} \sum_{m=1}^{M} a_{s(kp)} \cdot \alpha_{sm} + K \cdot C = \sum_{k=1}^{K} \mathrm{TPP}(k) + \sum_{k=1}^{K} \sum_{p \in \mathrm{PP(k)}} \sum_{s \in S(kp)} \mathrm{TM}(s) + K \cdot C$ |
| No. of integer variables | $M \cdot C$ |
| No. of continuous variables | $C \sum_{k=1}^{K} \sum_{p \in \mathrm{PP(k)}} \sum_{m=1}^{M} a_{s(kp)}.\alpha_{sm} = C \sum_{k=1}^{K} \sum_{p \in \mathrm{PP(k)}} \sum_{s \in S(kp)} \mathrm{TM}(s)$ |
| No. of constraints | $2K + \sum_{k=1}^{K} \sum_{p \in \mathrm{PP(k)}} \mathrm{TS}(kp) + M \cdot C + 1 + C + 2C \sum_{k=1}^{K} \sum_{p \in \mathrm{PP(k)}} \sum_{m=1}^{M} a_{s(kp)} \cdot \alpha_{sm} + K \cdot C = 2K + \sum_{k=1}^{K} \sum_{p \in \mathrm{PP(k)}} \mathrm{TS}(kp) + M \cdot C + 1 + C + 2C \sum_{k=1}^{K} \sum_{p \in \mathrm{PP(k)}} \sum_{s \in S(kp)} \mathrm{TM}(s) + K.C.$ |

### 3.3.3 | *Formulation III: Minimisation of Amortised Cost of Machines Plus Annual Operating Costs*

This formulation is a modification to the model of Logendran et al. [23]. They suggested that amortised costs of machines be used as the objective function instead of investment costs and also that total annual operating costs be included in the objective function instead of being controlled by budgetary constraints. Their solution approach is hierarchical: the first phase focuses on determining the system's total machine requirement and selecting a unique process plan for each part. The second phase determines the part families and machine cells using any method. While problem-solving turns out to be relatively easier with a hierarchical approach, it may sometimes lead to a suboptimal solution in the subsequent stages. This aspect will be illustrated with the help of a numerical example. For their model, a modified formulation (formulation III) is proposed, which uses a simultaneous rather than hierarchical approach to make various decisions. In fact, the modified formulation is like Logendran's model regarding only the objective function because, as far as constraints are concerned, it is similar to formulation II.

#### 3.3.3.1 | *Objective Function.*

Minimisation of the total amortised cost of machines plus annual operating costs can be written as:

$$\min \left[ \sum_{c=1}^{C} \sum_{m=1}^{M} I_{am}.N_{mc} + \sum_{c=1}^{C} \sum_{k=1}^{K} \sum_{p \in \mathrm{PP(k)}} \sum_{s \in S(\mathrm{kp})} \sum_{m \in M(\mathrm{s})} d_k \cdot X_{s(kp)mc} \cdot \mathrm{OC}_{s(kp)m} \right]. \tag{13}$$

#### 3.3.3.2 | *Constraints.*

1. Selection of only one process plan for each part: Same as constraint (8)
2. Assignment of operations: Same as constraint (9)
3. Loads on machines: Same as constraint (4)
4. Number of machines assigned to a cell: Same as a constraint (5b)

### 3.3.4 | *Formulation IV: Minimisation of Intercell and Intracell Movements*

In both formulations II and III, constraint (9) keeps the total intercell movements at zero, thus resulting in the formation of disjoint cells. However, each machine may be required by many operations of different parts, so many copies of that machine must be acquired for different cells. This will make the investment cost very high. On the other hand, if one insists that only one copy (or the minimum number of copies according to the total load for that machine type) be acquired for each machine type, then though the investment will be at its lowest, intercell movement will be considerable. The GT philosophy does not favour any of these extreme approaches, recommending instead a trade-off between these two costs, the investment and the intercell movement. This trade-off can be achieved in two ways. First, the total investment cost in the objective function should be minimised while total intercell movement is subject to a judicious non-zero maximum value. Another, minimise total intercell movement by the objective function while total investment cost is subject to a judiciously determined budgetary limit.

Both these approaches demand a judicious value for maximum total intercell movement or maximum total investment allowed, respectively. This judgement will depend on many system factors and, more than anything else, on the experience of the modeller. Obviously, when the total number of machines has to be kept at a minimum, the machines have to be assigned to the most demanding cells. In such cases, intercellular movements are bound to occur. A formulation that adopts the second of the approaches given above is presented here, that is, minimise total intercell and intracellular movement while total investment costs are kept within budgetary limits.

*3.3.4.1* | *Objective Function.* Same as the objective function (1).

*3.3.4.2* | *Constraints.*

1. Selection of only one process plan for each part: Same as constraint (2)
2. Assignment of operations: Same as constraint (3)
3. Total operating cost: Same as constraint (10)
4. Loads on machines: Same as constraint (11)
5. Number of machines assigned to a cell: Same as a constraint (5b)
6. Total investment in machines: $\sum_{c=1}^{C} \sum_{m=1}^{M} I_m \cdot N_{mc} \leq B$

## 4 | Numerical Illustration and Analysis of Results

### 4.1 | Formulations I, II, III and V (Process Plans are Given)

The problem described by Rajamani et al. [22] is used to illustrate the workings of all the formulations. The problem is as follows. Four different part types with known demand, each having multiple process plans as given in Table 1, are to be manufactured. Each operation in a process plan can be performed on more than one alternative machine. Three types of machines with known capacities and investment costs are available and their compatibility with various operations is given in Table 2. The time and cost information for performing various operations on different machines is given in Table 3.

Formulation I (grouping at the operational level: minimisation of intercell and intracell movement) was applied to the data given in Tables 1–3, with the additional information that the number of machines available for each type is one, the maximum number of machines that can be assigned to any cell is two and number of cells to be formed is two. Tables 4a and 4b show the results.

From Table 4(a), it is evident that there will be two intercell movements, one for part 1 and one for part 2 and two intracell movements, one for part 3 and one for part 4.

Formulation II (grouping at design level: minimisation of total investment cost, disjoint cells are to be formed), which is a modification to the model of Rajamani et al. [22], was also used to solve the example problem given in Tables 1–3, with the additional information that number of cells to be formed is two and the maximum number of machines that can be assigned to any cell is two. Tables 5a and 5b show the same results obtained by the model of [22].

However, the number of variables and constraints involved in both the formulations is different, as shown below:

| | Formulation III | Model of Rajamani et al. [22] |
|---|---|---|
| No. of binary variables | 102 | 59 |
| No. of integer variables | 6 | 6 |
| No. of continuous variables | 0 | 84 |
| No. of constraints | 55 | 214 |

Formulation III (grouping at design level: minimisation of amortised cost and annual operating costs; disjoint cells are to be formed), which is a modification to the model of Logendran et al. [23], was used for the same example problem (Tables 1–3) with one modification that the investment cost ($I_{\mathrm{m}}$) is replaced by amortised cost ($I_{\mathrm{am}}$). All other information is the same as that used for solving formulation II. The results obtained are the same as those given by formulation II, with the amortised cost being 600. This is not surprizing because, as we have observed before, formulation III is very similar to formulation II. The annual operating costs, another part of the objective function, equal 330, making the objective function 930. The two costs for

**TABLE 2** | Operation-machine compatibility.

| Operation | Machine type $m = 1$ | $m = 2$ | $m = 3$ |
|---|---|---|---|
| $s = 1$ | 1[a] | — | 1 |
| $s = 2$ | — | 1 | 1 |
| $s = 3$ | 1 | 1 | — |
| Capacity($b_m$) | 100 | 100 | 100 |
| Investment cost ($I_m$) | 100 | 250 | 300 |

[a]Entry '1' indicates that operation $s$ can be performed by machine type $m$ and '—' indicates that it is not possible.

**TABLE 1** | Part-process plan-operations incidence matrix.

| | Part | | | | | | | | |
|---|---|---|---|---|---|---|---|---|---|
| | $k = 1$ | | $k = 2$ | | $k = 3$ | | | $k = 4$ | |
| | Process plan | | | | | | | | |
| Operation | $p = 1$ | $p = 2$ | $p = 1$ | $p = 2$ | $p = 1$ | $p = 2$ | $p = 3$ | $p = 1$ | $p = 2$ |
| $s = 1$ | 1[a] | — | 1 | — | 1 | — | 1 | 1 | 1 |
| $s = 2$ | 1 | 1 | 1 | 1 | 1 | 1 | 1 | 1 | 1 |
| $s = 3$ | — | 1 | 1 | 1 | — | 1 | 1 | 1 | — |
| Demand ($d_k$) | 10 | | 10 | | 10 | | | 10 | |

[a]Entry '1' indicates that process plan $p$ of part $k$ requires operation $s$ and '—' indicates that operation $s$ is not required.

**TABLE 3** | Processing times $t_{s(kp)m}$ and costs $C_{s(kp)m}$ for operations.

| | | Part | | | | | | | | |
|---|---|---|---|---|---|---|---|---|---|---|
| | | $k = 1$ | | $k = 2$ | | $k = 3$ | | | $k = 4$ | |
| | | Process plan | | | | | | | | |
| **Operation** | **machine** | $p = 1$ | $p = 2$ | $p = 1$ | $p = 2$ | $p = 1$ | $p = 2$ | $p = 3$ | $p = 1$ | $p = 2$ |
| $s = 1$ | $m = 1$ | 5[a], 3[b] | — | 3, 4 | — | 2, 2 | — | 8, 1 | 1, 2 | 9, 7 |
| | $m = 3$ | 7, 2 | — | 4, 3 | — | 2, 2 | — | 9, 2 | 2, 1 | 8, 9 |
| $s = 2$ | $m = 2$ | 3, 5 | 9, 8 | 7, 8 | 3, 3 | 3, 3 | 1, 2 | 5, 9 | 2, 3 | 9, 8 |
| | $m = 3$ | 4, 3 | 7, 9 | 7, 7 | 2, 3 | 4, 4 | 2, 4 | 3, 10 | 2, 4 | 10, 9 |
| $s = 3$ | $m = 1$ | — | 8, 8 | 10, 9 | 6, 5 | — | 11, 7 | 7, 4 | 3, 5 | — |
| | $m = 2$ | — | 7, 7 | 8, 9 | 6, 6 | — | 8, 8 | 9, 5 | 2, 6 | — |

[a,b]Processing time and cost for doing operation $s$ of process plan $p$ of part $k$ on machine type $m$.

**TABLE 4a** | Process plan selection and assignment of operations to machines in different cells.

| | Part | | | |
|---|---|---|---|---|
| | $k = 1$ | $k = 2$ | $k = 3$ | $k = 4$ |
| | Process plan | | | |
| **Operation** | $p = 1$ | $p = 2$ | $p = 1$ | $p = 1$ |
| $s = 1$ | 3[a], 1[b] | — | 1, 2 | 1, 2 |
| $s = 2$ | 2, 2 | 3, 1 | 2, 2 | 2, 2 |
| $s = 3$ | — | 1, 2 | — | 2, 2 |

[a,b]For part $k$, the process plan selected is $p$ and operation $s$ is assigned to machine $a$ in cell $b$.

**TABLE 4b** | Assignment of machines to cells.

| | Cell | |
|---|---|---|
| **Machine type** | $c = 1$ | $c = 2$ |
| $m = 1$ | 1[a] | 0 |
| $m = 2$ | 1 | 0 |
| $m = 3$ | 0 | 1 |

[a]Entry ‘1’ indicates that 1 copy of machine type $m$ is assigned to cell $c$.

**TABLE 5a** | Process plan selection and assignment of operations to machines in different cells.

| **Objective function value = 600** | | | | |
|---|---|---|---|---|
| | Part | | | |
| | $k = 1$ | $k = 2$ | $k = 3$ | $k = 4$ |
| | Process plan | | | |
| **Operation** | $p = 1$ | $p = 2$ | $p = 1$ | $p = 1$ |
| $s = 1$ | 3[a], 1[b] | — | 1, 2 | 1, 2 |
| $s = 2$ | 2, 2 | 3, 1 | 2, 2 | 2, 2 |
| $s = 3$ | — | 1, 2 | — | 2, 2 |

[a,b]For part $k$, the process plan selected is $p$ and operation $s$ is assigned to machine $a$ in cell $b$.

the same problem, as solved by Logendran et al.’s model, are 550 and 350, respectively, summing up to 900. This solution (by Logendran et al.’s model) is shown in Table 6.

It is obvious from this solution (Table 6) that disjoint cells are not possible if the number of groups to be formed is more than one. If the number of groups to be formed is two and the maximum number of machines that can be assigned to any cell is two, then the best solution that can be obtained with this first phase solution is as follows. There will be two intercellular movements and three intracellular movements. On the other hand, two disjoint cells can be formed with three intracell movements by formulations II and III, with a total cost of only slightly more. Thus, this example clearly brings out the disadvantages of the hierarchical approach.

**TABLE 5b** | Assignment of machines to cells.

| | Cell | |
|---|---|---|
| **Machine type** | $c = 1$ | $c = 2$ |
| $m = 1$ | 1[a] | 0 |
| $m = 2$ | 1 | 0 |
| $m = 3$ | 0 | 1 |

[a]Entry ‘1’ indicates that 1 copy of machine type $m$ is assigned to cell $c$.

**TABLE 6** | Process plan selection and assignment of operations to machines (first phase).

| **Objective function value = 900** | | | | |
|---|---|---|---|---|
| | Part | | | |
| | $k = 1$ | $k = 2$ | $k = 3$ | $k = 4$ |
| | Process plan | | | |
| **Operation** | $p = 1$ | $p = 2$ | $p = 2$ | $p = 1$ |
| $s = 1$ | 1[a] | — | — | 1 |
| $s = 2$ | 2 | 2 | 2 | 2 |
| $s = 3$ | — | 1 | 1 | 1 |

[a]Entry ‘1’ indicates that for part $k$, the process plan selected is $p$ and operation $s$ is assigned to machine type $a$.

Formulation IV (grouping at design level: minimisation of intercell and intracell movement) was also used to solve the example problem given in Tables 1–3, with the additional information that the number of cells to be formed is two and the maximum number of machines that can be assigned to any cell is two. Here, an interesting difference is observed that when the budget limit for investment on machines is kept at 600, the solution is exactly the same as obtained by formulation II (Tables 5a and 5b), but when the budget restriction on machines is put at 550 the results are changed and Tables 7a and 7b shows the results.

**TABLE 7a** | Process plan selection and assignment of operations to machines in different cells.

| Objective function value = 3 | | | | |
|---|---|---|---|---|
| | Part | | | |
| | $k = 1$ | $k = 2$ | $k = 3$ | $k = 4$ |
| | Process plan | | | |
| **Operation** | $p = 1$ | $p = 2$ | $p = 2$ | $p = 1$ |
| $s = 1$ | 1[a], 1[b] | — | — | 1, 1 |
| $s = 2$ | 2, 1 | 2, 1 | 2, 1 | 2, 1 |
| $s = 3$ | — | 1, 2 | 1, 2 | 1, 1 |

[a,b]For part $k$, the process plan selected is $p$ and operation $s$ is assigned to machine $a$ in cell $b$.

**TABLE 7b** | Assignment of machines to cells.

| | Cell | |
|---|---|---|
| **Machine type** | $c = 1$ | $c = 2$ |
| $m = 1$ | 1[a] | 2 |
| $m = 2$ | 1 | 0 |
| $m = 3$ | 0 | 0 |

[a]Entry '1' indicates that 1 copy of machine type $m$ is assigned to cell $c$.

From the results (Tables 5a, 5b, 7a and 7b), it is obvious that formulation IV is more flexible than formulation II in terms of managerial decision-making. When there is no budget limit on machine investment, formulation IV will always produce the same results as formulation II in intercell movements and perhaps better results in intracell movements.

The proposed integer programming formulations were implemented using IBM ILOG CPLEX Optimization Studio (version 22.1). All computational experiments were executed on a workstation equipped with an Intel Core i7 processor and 16 GB RAM under the Windows operating system. Default parameter settings of CPLEX were used unless otherwise specified.

## 4.2 | Practical Relevance and Managerial Implications

The four mathematical formulations developed in this study address different decision-making contexts encountered in the design and operation of cellular manufacturing systems. In real manufacturing environments, the selection of an appropriate formulation depends on the strategic priorities of the production system, such as minimising movement costs, reducing investment expenditure or balancing operational efficiency with budget constraints.

### *4.2.1 | Application of Formulation I: Operational Reorganization*

Formulation I is particularly suitable when an existing manufacturing system needs to be reorganised without acquiring new machines. In such situations, the primary objective of management is to improve production flow by minimising intercell and intracell movements while utilising the currently available machines. In mature manufacturing environments, such as a Small-to-Medium Enterprise (SME) producing a variety of automotive spare parts (e.g., gears, shafts and housings), the machine fleet is often already established. When market demand shifts or new product variants are introduced, managers must optimise the shop floor layout without additional capital expenditure. In such a scenario, Formula I is most appropriate. It allows plant managers to reorganise existing CNC lathes, milling machines and grinders into cells that minimise material handling movements, thereby reducing lead times and work-in-process (WIP) inventory using current resources.

### *4.2.2 | Application of Formulation II: Cost-Oriented System Design*

Formulation II is applicable when a new cellular manufacturing system is being designed and the primary managerial objective is to minimise the total investment cost in machines while ensuring that each part family is processed within a single cell. This formulation is relevant for organisations planning to establish new production facilities with strict capital budget constraints. For instance, when a manufacturer plans to introduce a new product line requiring several machining operations, different machine types must be acquired. In such a scenario, the formulation determines the minimum number of machines required across cells while ensuring that all operations of a part are completed within a single cell, thereby eliminating intercell movements. The resulting decision variables $N_{mc}$ determine the number of machines of type $m$ assigned to cell $c$ and the objective function: $\min \sum_{c=1}^{C} \sum_{m=1}^{M} I_m N_{mc}$, ensures that the total investment cost remains minimal.

### *4.2.3 | Application of Formulation III: Long-Term Cost Optimization*

Formulation III extends the previous model by minimising the sum of amortised machine costs and annual operating costs. This formulation is particularly relevant for long-term strategic planning, where managers must consider both capital expenditure and operational efficiency. For example, in an automotive component manufacturing plant producing gears and transmission components, machines such as CNC machining centres and grinding machines involve significant acquisition costs and operating expenses. By incorporating amortised machine costs and operating costs into the objective function, Formulation III allows decision-makers to identify machine-cell configurations that minimise the overall lifecycle cost of the manufacturing system.

### *4.2.4 | Application of Formulation IV: Budget-Constrained Movement Minimisation*

Formulation IV is most suitable when organisations aim to reduce intercell and intracell movements but must operate under a fixed investment budget. In such cases, management specifies an upper limit on total investment and allows the optimization model to determine the machine-cell configuration that minimises material handling movements. For example, suppose a factory modernisation project has a predetermined investment budget $B$. The objective then becomes minimising movement-related inefficiencies while satisfying the investment constraint: $\sum_{c=1}^{C} \sum_{m=1}^{M} I_m N_{mc} \leq B$. This formulation enables managers to balance the trade-off between material handling efficiency and capital investment, which is a common practical challenge in manufacturing system design.

In summary, the proposed formulations provide a flexible decision-support framework that can be adapted to different industrial contexts. Depending on whether the managerial priority is operational improvement, investment minimisation, lifecycle cost optimization or movement reduction under budget constraints, the appropriate formulation can be selected and applied to guide the design and operation of cellular manufacturing systems.

## 5 | Conclusions

In this paper, an effort has been made to develop mathematical programming formulations for the generalised grouping problem at the design and operational levels for a cellular manufacturing system. These formulations will help understand and analyse a problem in its entirety, although when the problem size is large, solving these may not be easy. These formulations may also provide valuable clues for designing heuristics, comparing solutions obtained by a heuristic and comparing the performance of different heuristics. As a summary of the paper, briefly describing all the formulations developed would be pertinent.

Formulation I seek to minimise intercell and intracell movement by simultaneously grouping the machines and assigning the consecutive operations of selected process plans to the same cell and to the same machine. It can be used to reorganise an existing CMS into a new CMS. On the other hand, formulations II, III and IV have been developed in this paper to set up a new CMS. Formulation II minimises the total investment costs on machines, while formulation III minimises the total amortised costs of machines plus annual operating costs. Formulations II and III will always produce disjoint groups. To overcome the shortcomings of formulations II and III, formulation IV is proposed, which minimises intercell and intracell movements while keeping the investment in machines within the budgetary limit. Formulation IV is found to be more flexible than formulations II and III in terms of managerial decision-making.

The models presented in this paper explicitly take into account the limitations on the number of machines in a group and the capacity of machines of a particular type. They also incorporate some of the economic factors that are bound to affect grouping decisions. All the models developed in this paper are NP-hard, as stated by Logendran et al. [23] and become computationally expensive as the problem size increases in terms of the number of parts, operations, process plans and machines. The major recommendation for further research could be the application of modern heuristic methods like tabu search, simulated annealing and genetic algorithms to solve these problems, as these techniques have been found to be effective in solving combinatorial optimization problems.

### Author Contributions

**Md. Kutub Uddin:** conceptualization, formal analysis, methodology, software, writing – original draft. **Md. Saiful Islam:** formal analysis, methodology, validation, writing – original draft. **Md Abrar Jahin:** formal analysis, software, validation, writing – original draft. **Md. Tanjid Hossen Irfan:** data curation, formal analysis, visualization, writing – original draft. **Md. Saiful Islam Seam:** data curation, formal analysis, visualization, writing – original draft. **M. F. Mridha:** investigation, validation, writing – review and editing.

### Funding

The authors have nothing to report.

### Ethics Statement

This study did not involve any human participants, animals or personally identifiable data requiring ethics approval or consent to participate.

### Consent

All authors have read and approved the final manuscript and consent to its publication.

### Conflicts of Interest

The authors declare no conflicts of interest.

### Data Availability Statement

The datasets and models generated or analyzed during the study are available from the corresponding author upon reasonable request.